\pdfoutput=1

\documentclass[11pt]{article}

\usepackage[final]{acl}
\usepackage{graphicx}

\usepackage{times}
\usepackage{latexsym}

\usepackage[T1]{fontenc}

\usepackage[utf8]{inputenc}

\usepackage{microtype}

\usepackage{inconsolata}

\usepackage[normalem]{ulem}
\usepackage{algorithm}
\usepackage{multirow}
\usepackage{amsmath} 
\usepackage{enumitem}

\usepackage{amsmath}
\usepackage{booktabs}

\usepackage{algorithm}
\usepackage{algpseudocode}
\usepackage[most]{tcolorbox}

\title{DRAGIN: Dynamic Retrieval Augmented Generation based on the Information Needs of Large Language Models}

\usepackage{authblk}
\author[1]{\textbf{Weihang Su}\thanks{swh22@mails.tsinghua.edu.cn}}
\author[1]{\textbf{Yichen Tang}\thanks{contributed equally}}
\author[1]{\textbf{Qingyao Ai}\thanks{Corresponding Author: aiqy@tsinghua.edu.cn}}
\author[2]{\textbf{Zhijing Wu}}
\author[1]{\textbf{Yiqun Liu}}

\affil[1]{Department of Computer Science and Technology, Tsinghua University}
\affil[2]{School of Computer Science and Technology, Beijing Institute of Technology}

\begin{document}
\maketitle

\begin{abstract}

Dynamic retrieval augmented generation (RAG) paradigm actively decides when and what to retrieve during the text generation process of Large Language Models (LLMs).
There are two key elements of this paradigm: identifying the optimal moment to activate the retrieval module (deciding when to retrieve) and crafting the appropriate query once retrieval is triggered (determining what to retrieve).
However, current dynamic RAG methods fall short in both aspects. 
Firstly, the strategies for deciding when to retrieve often rely on static rules. 
Moreover, the strategies for deciding what to retrieve typically limit themselves to the LLM's most recent sentence or the last few tokens, while the LLM's information needs may span across the entire context.
To overcome these limitations, we introduce a new framework, DRAGIN, i.e., \underline{D}ynamic \underline{R}etrieval \underline{A}ugmented \underline{G}eneration based on the \underline{I}nformation \underline{N}eeds of LLMs. 
Our framework is specifically designed to make decisions on when and what to retrieve based on the LLM's information needs during the text generation process.
We evaluate DRAGIN along with existing methods comprehensively over 4 knowledge-intensive generation datasets. Experimental results show that DRAGIN achieves superior performance on all tasks, demonstrating the effectiveness of our method\footnote{We have open-sourced all the code, data, and models in GitHub: https://github.com/oneal2000/DRAGIN/tree/main}.

\end{abstract}


\section{Introduction}

In recent years, large language models (LLMs) have made significant advancements across various natural language processing (NLP) tasks, quickly becoming a critical element in numerous AI applications~\cite{brown2020language,chowdhery2022palm,touvron2023llama,scao2022bloom,zhang2022opt}. Despite their impressive capabilities, these models often produce text that seems coherent and plausible but factually incorrect, a problem commonly known as hallucination~\cite{maynez2020faithfulness,zhou2020detecting,liu2021token,ji2023survey,su2024unsupervised}.

To mitigate this issue, Retrieval-Augmented Generation (RAG) has emerged as a prominent solution. 
RAG enhances LLMs by retrieving and incorporating relevant information from external databases into the LLMs' inputs. 
It has demonstrated superior effectiveness across numerous NLP challenges~\cite{khandelwal2019generalization,borgeaud2022improving,lewis2020retrieval,guu2020retrieval,izacard2020leveraging,jiang2022retrieval,shi2023replug}. 
Traditional methods of RAG typically rely on single-round retrieval, using the LLM's initial input to retrieve relevant information from external corpora. 
While this method is effective for straightforward tasks, it tends to fall short for complex multi-step tasks and long-form generation tasks~\cite{jiang2023active}. 
In contrast, dynamic RAG~\cite{trivedi2022interleaving,borgeaud2022improving,ram2023context,jiang2023active} performs multiple times of retrieval during the generation process of LLMs. 
It includes two steps: identifying the optimal moment to activate the retrieval module (deciding when to retrieve), and crafting the appropriate query once retrieval is triggered (determining what to retrieve).
Depending on when and what to retrieve, a variety types of methods have been proposed in this direction.
For example, IRCoT~\cite{trivedi2022interleaving} adopts a global augmentation method where retrieval is conducted for each generated sentence, with the latest generated sentence used as the query. 
RETRO~\cite{borgeaud2022improving} and IC-RALM~\cite{ram2023context} define a sliding window and trigger the retrieval module based on a preset number of processed tokens, and the last $n$ tokens are used as the query.

However, existing dynamic RAG methods face several critical challenges, primarily in determining the optimal timing for retrieval and formulating effective queries when retrieval is triggered. 
First of all, existing approaches often rely on static rules to decide when to retrieve, neglecting the assessment of necessity and potential risks involved.
On the one hand, depending on the quality of the input query and retrieval models, unnecessary retrieval augmentation may introduce irrelevant or noisy data to LLMs which could jeopardize the quality of the outputs. 
On the other hand, conducting retrieval augmentation will inevitably increase the time and computation cost of LLM inference, such cost is unworthy if LLMs can generate correct outputs by themselves.
Additionally, the strategies of existing studies in determining what to retrieve often restrict themselves to the LLM's most recent sentence or the last few tokens. 
This approach may not capture the model's real-time information needs since the LLM's information needs may actually be related to terms that span the entire context. Retrieving documents in this manner is thus suboptimal in many cases.


To overcome these limitations, we introduce a new framework, \textbf{DRAGIN}, i.e., \underline{D}ynamic \underline{R}etrieval \underline{A}ugmented \underline{G}eneration based on the \underline{I}nformation \underline{N}eeds of LLMs. Our framework is specifically designed to make decisions on when and what to retrieve, based on the LLM's information needs during the text generation process.
For the timing of retrieval, we propose \textbf{RIND}: \underline{R}eal-time \underline{I}nformation \underline{N}eeds \underline{D}etection, which considers the LLM's uncertainty about its own generated content, the influence of each token on subsequent tokens, and the semantic significance of each token.
For the formulation of retrieval queries, we propose \textbf{QFS}: \underline{Q}uery \underline{F}ormulation based on \underline{S}elf-attention, which innovates query formulation by leveraging the LLM's self-attention across the entire context.
DRAGIN is a lightweight RAG framework that can be incorporated into any Transformer-based LLMs without further training, fine-tuning, or prompt engineering.

We comprehensively evaluate DRAGIN along with existing dynamic RAG frameworks over four knowledge-intensive generation benchmarks. Experimental results show that DRAGIN achieves superior performance on all datasets, demonstrating the effectiveness of our method. 
Moreover, the results of the ablation study indicate that our proposed new strategies for "when to retrieval" (i.e., RIND) and "what to retrieval" (i.e., QFS) perform uniformly better than other strategies in existing RAG methods despite retrieval models and LLMs. 

In summary, the contributions of our paper are as follows:

\begin{itemize}[leftmargin=*]
\item We propose a novel dynamic RAG framework: DRAGIN. In contrast to previous works, our framework optimizes when and what to retrieve based on the real-time information needs of the LLM.

\item We evaluate existing dynamic RAG methods and DRAGIN on four knowledge-intensive datasets using three different LLMs. Experimental results indicate that DRAGIN achieves state-of-the-art (SOTA) performance.

\end{itemize}

\section{Related Work}
\subsection{Single-round Retrieval-augmented LLM}

LLMs have demonstrated significant effectiveness across a wide range of tasks. 
However, their built-in knowledge can sometimes fall short when dealing with knowledge-intensive tasks. 
To address this limitation, Retrieval-Augmented Generation (RAG) strategies are widely employed to enhance the performance of LLMs.
One of the most direct methods is single-round retrieval augmentation~\cite{khandelwal2019generalization,borgeaud2022improving,lewis2020retrieval,guu2020retrieval,izacard2020leveraging,jiang2022retrieval,shi2023replug}, which involves using the initial input as a query to retrieve information from an external corpus. 
The retrieved external knowledge is then incorporated as part of the input for the model. 
Previous research has explored single-round retrieval augmentation extensively. 
For instance, REPLUG~\cite{shi2023replug} treats LLMs as a black box and leverages them to generate training data for the retrieval model. 
From a different perspective, UniWeb~\cite{li2023web} proposes an adaptive search engine-assisted learning method that can self-assess whether the LLM requires retrieval augmentation.

\subsection{Multi-round Retrieval-augmented LLM}

Single-round retrieval can be relatively effective for simple tasks or cases where user information needs are clear-cut. 
However, for complex tasks or tasks involving the generation of lengthy text, such as long-form question answering, multi-hop reasoning, chain-of-thought reasoning, etc., relying solely on the user's initial input for retrieval may not adequately cover all the external knowledge that the model requires~\cite{jiang2023active}. 
Therefore, some researchers have begun to explore multi-round retrieval augmentation. For example, RETRO~\cite{borgeaud2022improving} and IC-RALM~\cite{ram2023context} trigger retrieval every 4 to 32 tokens, and IRCot~\cite{trivedi2022interleaving} triggers retrieval every sentence. 
However, solely relying on fixed interval-based retrieval without considering the information needs of the LLM itself could produce suboptimal results. 
Inspired by this, FLARE~\cite{jiang2023active} triggers retrieval when encountering an uncertain token.
Specifically, if any token in the generated text has a probability lower than a certain threshold, the retrieval module is triggered. 


\begin{figure}[t]
\centering
    \includegraphics[width=\columnwidth]{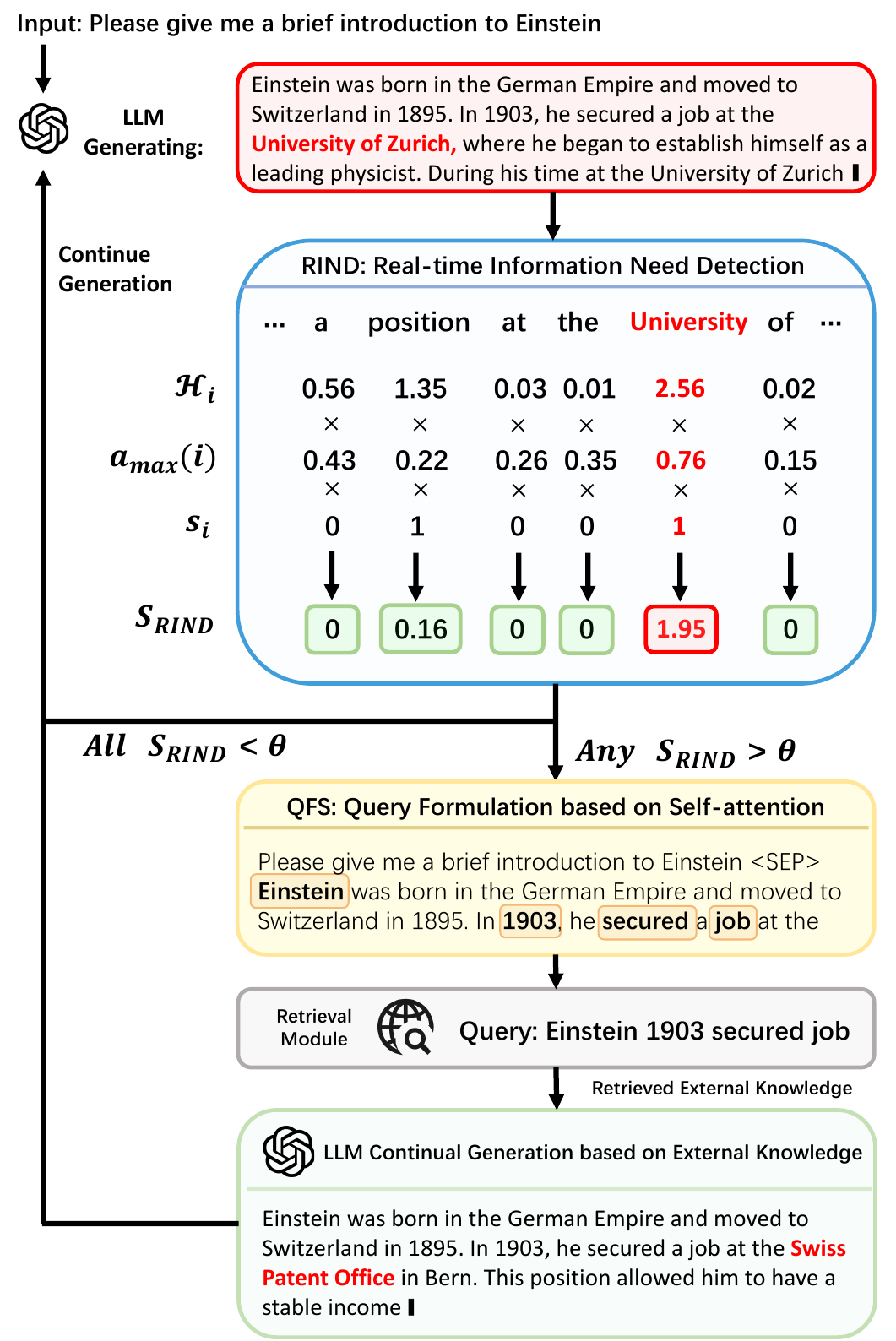}
    \small{\caption{An illustration of our DRAGIN framework.} \label{pic:framework}}
\end{figure}

\section{Methodology}
In this section, we introduce the DRAGIN framework in detail. DRAGIN consists of two components: Real-time Information Needs Detection (RIND) and Query Formulation based on Self-attention (QFS), as illustrated in Figure~\ref{pic:framework}. We introduce RIND in section~\ref{sec:RIND}, and QFS in section~\ref{sec:QFS}.

\subsection{Real-time Information Need Detection}
\label{sec:RIND}

As discussed above, most existing dynamic RAG frameworks trigger the retrieval module based on static, predefined rules. 
To the best of our knowledge, the only notable exception is FLARE~\cite{jiang2023active} which triggers retrieval dynamically when the LLM's confidence (i.e., the generation probability) on the next token is lower than certain thresholds. 

However, the necessity of retrieval augmentation not only depends on the generation confidence, but also depends on the importance of the token, the semantic  of the token, and the influence of each token on subsequent tokens.
To address the limitations of the existing approaches, we propose an enhanced approach for triggering retrieval within dynamic RAG frameworks, named Real-time Information Needs Detection (RIND). 
This method refines the retrieval activation process by evaluating not only the uncertainty of each token, but also its semantic contribution and the impact on the following context.

RIND begins by quantifying the uncertainty of each token generated during the LLM's inference process. This is accomplished by recording the entropy of the token's probability distribution across the vocabulary. 
Consider an output sequence generated by an LLM, denoted as \( T = \{t_1, t_2, \ldots, t_n\} \), with each \( t_i \) representing an individual token within the sequence at position $i$.
For any token \(t_i\), the entropy \(\mathcal{H}_i\) is computed as follows:
\begin{equation}
\small
    \mathcal{H}_i = -\sum_{{v} \in \mathcal{V}} p_i({v}) \log p_i({v}),
\end{equation}
where $p_i(v)$ denotes the probability of generating the token $v$ over all tokens in the vocabulary $\mathcal{V}$ at position $i$.
This measurement of uncertainty serves as the first dimension in our multi-faceted evaluation of tokens.

In addition, RIND leverages the self-attention mechanism inherent in Transformer-based LLMs to allocate weights to tokens, which represent the tokens' impact on the subsequent context.
Specifically, for any given token \(t_i\), we quantify its influence by recording the maximum attention value \(a_{\max}(i)\), which records the maximum attention from all following tokens~\footnote{We choose the attention scores of the last Transformer layer of the LLM.}.  
The attention matrix \(A\) is computed as follows:
\begin{equation}
\small
A = \text{softmax}\left(\text{mask}\left(\frac{QK^\top}{\sqrt{d_k}}\right)\right),
\end{equation}
where \(Q\) is the query matrix, \(K\) is the key matrix, \(d_k\) denotes the dimensionality of a key vector. 
The mask function is applied to the scaled dot-product matrix \( M = \frac{QK^\top}{\sqrt{d_k}} \) to prevent each position from attending to subsequent positions. Specifically, for the square matrix \( M \) (with dimensions equal to the sequence length), we set \( M_{i,j} = -\infty \) whenever \( i < j \). This effectively masks out the upper triangular part of the matrix. The softmax function is then applied row-wise to produce the attention weights.

Following this, the maximum attention value \(a_{\max}(i)\) for token \(t_i\) is identified by locating the highest \(A_{j,i}\) for all \(j > i\):
\begin{equation}
\small
a_{\max}(i) = \max_{j > i} A_{j,i}
\end{equation}



Consider the semantic contribution of each token, RIND employs a binary semantic indicator to filter out stopwords, thus concentrating on tokens with significant semantic value:
\begin{equation}
\small
 s_i = \begin{cases} 0, & \text{if } t_i \in S \\ 1, & \text{otherwise} \end{cases} ,
\end{equation}
where $S$ is the stopwords set, $s_i$ is the semantic contribution score of the token $t_i$. This process ensures that only semantically potent tokens contribute to the retrieval decision-making process.

Combining uncertainty, significance, and semantics, RIND computes a comprehensive score for each token $t_i$:
\begin{equation}
\small
\mathcal{S}_{RIND}(t_i) = \mathcal{H}_i \cdot a_{\max}(i) \cdot s_i
\end{equation}
Let $T = \{t_1, t_2, \ldots, t_n\}$ represent the set of tokens already generated by the LLM. The retrieval module activates when the score $\mathcal{S}_{RIND}(t_i)$ for any token exceeds a predefined threshold, \(\theta\).

\subsection{Query Formulation based on Self-attention}
\label{sec:QFS}

Once the position to conduct retrieval augmentation is determined, the next step in the RAG framework is to formulate a query to retrieve necessary information from external databases for the continued generation of LLMs.
In the existing dynamic RAG frameworks, all the query formulation methods limit their focus to the LLM's most recent sentence or the last few tokens. This narrow scope fails to adequately cater to the model's real-time information needs, which may span across the entire context. To overcome the shortcomings of these approaches, we propose a novel strategy that utilizes the self-attention mechanisms inherent in Transformer-based LLMs. Our method, termed "Query Formulation based on Self-Attention" (QFS), seeks to ascertain the LLM's information needs more precisely by examining its understanding of the full context.

Consider an output sequence generated by an LLM, denoted as \( T = \{t_1, t_2, \ldots, t_n\} \), with each \( t_i \) representing an individual token within the sequence. Suppose the RIND module identifies the token at position \( i \), which requires external knowledge and triggers the retrieval module. 
The QFS approach then focuses on this specific position to formulate a query. 
For the token at position \( i \), the QFS method evaluates the attention weights across the preceding token sequence \( \{t_{i-1}, t_{i-2}, ..., t_1\} \). 
Since the generation of \( t_i \) by the LLM is based on its interpretation of the entire preceding context, the attention weights reflect the model's self-assessed importance of each token in generating \( t_i \). 
The QFS method prioritizes these tokens based on their attention scores, selecting the top \( n \) tokens to construct the query. 
The query formulation process includes the following steps:
(1) Extract the attention scores of the last Transformer layer \( A_i = \{a_{i,1}, a_{i,2}, ..., a_{i,i-1}\} \) for each token \( t_i \) in \( T \), where \( a_{i,j} \) represents the attention score assigned by \( t_i \) to \( t_j \); 
(2) Sort \( A_i \) in descending order to identify the top \( n \) tokens with the highest attention scores;
(3) Find the words corresponding to these tokens from the vocabulary and arrange them according to their original order in the text;
(4) Construct the query \( Q_i \) using the words from these top \( n \) tokens, ensuring the query reflects the most relevant aspects of the context as determined by the LLM's self-attention mechanism.

\subsection{Continue Generation after Retrieval}

Once the RIND module detects the position $i$ that needs external knowledge, the QFS module creates the query and utilizes an off-the-shelf retrieval model (e.g. BM25) to retrieve relevant information from external knowledge bases. 
Suppose the retrieved documents are denoted as $Di_1$, $Di_2$, and $Di_3$.
Upon successful retrieval, the next step of the dynamic RAG framework is to integrate this external knowledge into the LLM's generation process. This integration begins with truncating the LLM's output at the identified position $i$ for retrieval augmentation:
\begin{equation}
\small
T' = \text{truncate}(T, t_{i}),
\end{equation}
where \(T'\) represents the truncated output, \(T\) is the original sequence generated by the LLM, and \(t_{i}\) is the token at which the need for external knowledge was identified by RIND. 
To integrate the retrieved knowledge $Di_1$, $Di_2$, and $Di_3$, we adopt a meticulously designed prompt template~\footnote{The specific content of the prompt template is presented in Appendix~\ref{appendix:prompt}.}, which is structured as follows:
\begin{tcolorbox}[colback=lightgray!20,colframe=darkgray!80,title=The entire input for LLM:]
Below are the external knowledge references:

[1] \textbf{$Di_1$}

[2] \textbf{$Di_2$}

[3] \textbf{$Di_3$}

Please answer the question based on the external knowledge:

Question: xxx

Answer: \textbf{T'}
\end{tcolorbox}
At this point, the LLM continues generating content based on the retrieved external knowledge and the truncated output $T'$.
Following the integration of the retrieved knowledge, the LLM resumes generating content from the truncation point, enhanced with additional information. 
This procedure allows the LLM to bridge the previously identified knowledge gap, facilitating a more informed and precise continuation of its output.

Suppose at a subsequent position $j$ where RIND detects again that the LLM requires external knowledge. In that case, the QFS module is triggered again at position $j$ to generate a new query, retrieving a new set of documents $Dj_1$, $Dj_2$, and $Dj_3$ to replace $Di_1$, $Di_2$, and $Di_3$.
The LLM will then continue generating from position $j$ based on the newly retrieved documents, following the same process.

\section{Experimental Setup}

\subsection{Datasets}

We choose two MultihopQA datasets \textbf{2WikiMultihopQA}~\cite{ho2020constructing} and \textbf{HotpotQA}~\cite{yang2018hotpotqa} to evaluate the RAG framework's ability to answer complex questions that require multi-hop reasoning. We choose the \textbf{IIRC}~\cite{ferguson2020iirc} dataset to evaluate the RAG framework's ability in reading comprehension tasks. Furthermore, we utilize the \textbf{StrategyQA}~\cite{geva2021did} dataset to evaluate the commonsense reasoning capabilities of DRAGIN and other baselines.

\subsection{Settings for each Dataset}
\label{sec:setting}


\begin{itemize}[leftmargin=*]

    \item \textbf{2WikiMultihopQA}~\cite{ho2020constructing}. We follow the setting of~\cite{wang2022self} to generate both chain-of-thought (CoT) reasoning process as well as the final answer. We follow the prompt template of ~\cite{trivedi2022interleaving} and~\cite{jiang2023active}. 
    For the evaluation metrics, we extract the final answer from the generated output using pattern matching techniques. The extracted answer is then compared with the reference answer, utilizing methods such as exact match at the answer level, along with token-level measurements of F1 score and precision.

    \item \textbf{HotpotQA}~\cite{yang2018hotpotqa}. We follow the setting and the prompt template of~\cite{trivedi2022interleaving} to generate both chain-of-thought (CoT) reasoning process as well as the final answer. Our evaluation metric on this dataset is the same as 2WikiMultihopQA.



    \item \textbf{StrategyQA}~\cite{geva2021did}. We follow the setting of~\cite{wei2022chain} to generate both the CoT reasoning process as well as the final answer. We follow the prompt template of~\cite{wei2022chain} and~\cite{jiang2023active}. 
    For the evaluation metrics, the obtained yes/no response is extracted and compared with the standard correct answer using an exact match approach.

    \item \textbf{IIRC}~\cite{ferguson2020iirc}. We follow the setting and the prompt template of~\cite{trivedi2022interleaving} to generate the final answer. Our evaluation metric on this dataset is the same as 2WikiMultihopQA.
\end{itemize}

Besides the settings introduced in this section, the specific prompt templates corresponding to each dataset are presented in Appendix~\ref{appendix:prompt}.
Appendix~\ref{appendix:datasets} provides more detailed descriptions of each dataset's settings.




\subsection{Baselines}

\begin{table}[t]
\caption{A comparative overview of our selected Retrieval-Augmented Generation baselines.}
\label{tab:baselines}
\setlength\tabcolsep{2.5pt} 
\footnotesize
{
\begin{tabular}{lcc}
\toprule
              & \textbf{Timing for Retrieval}     & \textbf{Query Formulation}                                                                  \\
              \midrule
\textbf{SR-RAG}  & Before Generation             & Initial Input                                                                      \\
\midrule
\textbf{FL-RAG}  & Per $n$ Tokens            & Last Generated Tokens                                                                     \\
\midrule
\textbf{FS-RAG}  & Per Sentence            & Last Generated Sentence                                                                     \\
\midrule
\textbf{FLARE}  & \scriptsize{\begin{tabular}[c]{@{}c@{}}Any token's probability\\ below the threshold\end{tabular}} & \scriptsize{\begin{tabular}[c]{@{}c@{}}Last generated Sentence\\ exclude low-probability tokens\end{tabular}} \\
\midrule
\textbf{DRAGIN} & \scriptsize{\begin{tabular}[c]{@{}c@{}}  Generated token's  \\importance and uncertainty \end{tabular}} & \scriptsize{\begin{tabular}[c]{@{}c@{}}LLM's attention over \\ the entire context\end{tabular}}          \\
\toprule
\end{tabular}
}
\end{table}

We choose the following Text Generation baselines for comparison. 
Following the setting of FLARE~\cite{jiang2023active}, we implemented the existing multi-round RAG frameworks using the same settings, with the only variation being the timing of triggering retrieval (when to retrieve) and the query formulation method when the retrieval is triggered (what to retrieve).

\begin{itemize}[leftmargin=*]

\item  \textbf{wo-RAG}. LLM provides direct answers to questions without RAG.
\item  \textbf{SR-RAG} ({Single-round RAG}). Relevant passages are retrieved from an external corpus based on the initial question. The retrieved passages are then added into the LLM's input.
\item  \textbf{FL-RAG} ({F}ix {L}ength RAG)~~\cite{khandelwal2019generalization, borgeaud2022improving,ram2023context}. A multi-round retrieval augmentation method that triggers the retrieval module every $n$ tokens. The tokens generated in the previous token window are utilized as the query.
\item  \textbf{FS-RAG} ({F}ix {S}entence RAG)~~\cite{trivedi2022interleaving}. A multi-round retrieval augmentation method that triggers the retrieval module every sentence. The last generated sentence are utilized as the query.
\item \textbf{FLARE}~\cite{jiang2023active}. A multi-round retrieval augmentation method that triggers retrieval each time it encounters an uncertain token. When the retrieval module is triggered, the last generated sentence without the uncertain tokens are defines as the query.

\end{itemize}



To illustrate the differences between DRAGIN and other dynamic RAG baselines directly, we present a comparison of retrieval timing and query formation methods for each dynamic RAG frameworks in Table~\ref{tab:baselines}.

\begin{table*}[t]
\caption{The overall experimental results of DRAGIN and other baselines on four benchmarks. The best results are in bold.}
\label{tab:overall}
\centering
\small
\begin{tabular}{ccccccccc}
\toprule
                                          &                 & \multicolumn{2}{c}{\textbf{2WikiMultihopQA}} & \multicolumn{2}{c}{\textbf{HotpotQA}} & \textbf{StrategyQA} & \multicolumn{2}{c}{\textbf{IIRC}} \\
                                          \midrule
\textbf{LLM}                              & \textbf{RAG Method} & \textbf{EM}          & \textbf{F1}           & \textbf{EM}       & \textbf{F1}       & \textbf{Accuracy}         & \textbf{EM}     & \textbf{F1}     \\
\toprule
\multirow{6}{*}{\textbf{Llama2-13b-chat}} & \textbf{wo-RAG} & 0.187                & 0.2721                & 0.223             & 0.3097            & 0.650                & 0.168& 0.2039          \\
                                          & \textbf{SR-RAG}  & 0.245                & 0.3364                & 0.263             & 0.3706            & 0.654                & \textbf{0.196}& \textbf{0.2303}\\
                                          & \textbf{FL-RAG} & 0.217                & 0.3054                & 0.177             & 0.2682            & 0.648               & 0.155& 0.1875          \\
                                          & \textbf{FS-RAG} & 0.270                 & 0.3610                 & 0.267             & 0.3715            & 0.655               & 0.171& 0.2061          \\
                                                                                    & \textbf{FLARE}  & 0.224                & 0.3076                & 0.180              & 0.2756            & 0.655               & 0.138& 0.1667          \\

                                          & \textbf{DRAGIN (Ours)} & \textbf{0.304}& \textbf{0.3931}& \textbf{0.314}& \textbf{0.4238}& \textbf{0.689}& 0.185& 0.2221\\
                                          \toprule
\multirow{6}{*}{\textbf{Llama2-7b-chat}}  & \textbf{wo-RAG} & 0.146                & 0.2232                & 0.184             & 0.2745            & \textbf{0.659}      & 0.139& 0.1731          \\
                                          & \textbf{SR-RAG}  & 0.169                & 0.2549                & 0.164             & 0.2499            & 0.645               & 0.187& 0.2258          \\
                                          & \textbf{FL-RAG} & 0.112                & 0.1922                & 0.146             & 0.2107            & 0.635               & 0.172& 0.2023          \\
                                          & \textbf{FS-RAG} & 0.189                & 0.2652                & 0.214             & 0.3035            & 0.629               & 0.178& 0.2157          \\
                                                                                    & \textbf{FLARE}  & 0.143                & 0.2134                & 0.149             & 0.2208            & 0.627               & 0.136& 0.1644          \\
                                                                                    
                                          & \textbf{DRAGIN (Ours)} & \textbf{0.220}& \textbf{0.2926}& \textbf{0.232}& \textbf{0.3344}& 0.641               & \textbf{0.192}& \textbf{0.2336}\\
                                          \toprule
\multirow{6}{*}{\textbf{Vicuna-13b-v1.5}} & \textbf{wo-RAG} & 0.146                & 0.2232                & 0.228             & 0.3256            & 0.682               & 0.175& 0.2149          \\
                                          & \textbf{SR-RAG}  & 0.170                 & 0.2564                & 0.254             & 0.3531            & {0.686}      & 0.217           & 0.2564          \\
                                          & \textbf{FL-RAG} & 0.135                & 0.2133                & 0.187             & 0.3039            & 0.645               & 0.0985          & 0.1285          \\
                                          & \textbf{FS-RAG} & 0.188                & 0.2625                & 0.185             & 0.3216            & 0.622               & 0.1027          & 0.1344          \\
                                                                                    & \textbf{FLARE}  & 0.157                & 0.2257                & 0.092             & 0.1808            & 0.599               & 0.1174          & 0.1469          \\

                                          & \textbf{DRAGIN (Ours)} & \textbf{0.252}& \textbf{0.3516}& \textbf{0.288}& \textbf{0.4164}& \textbf{0.687}               & \textbf{0.2233}& \textbf{0.2652}\\
                                          \toprule
\end{tabular}
\end{table*}
\subsection{Selected LLMs}
To validate the effectiveness of DRAGIN and other RAG baselines, we conducted experiments with the following LLMs:

\begin{itemize}[leftmargin=*]
\item \textbf{LLaMA-2-Chat}~\cite{touvron2023llama2}. LLaMA2 is a collection of pre-trained and fine-tuned LLMs. This series includes fine-tuned LLMs, known as Llama2-Chat, specifically designed for optimal performance in dialogue-based applications. We choose \textbf{LLaMA-2-Chat-7B} and \textbf{LLaMA-2-Chat-13B}.


\item \textbf{Vicuna-13B-v1.5}~\cite{chiang2023vicuna} is a collection of open-source chatbots fine-tuned from LLaMA using user-shared conversations gathered from ShareGPT. We have selected the latest versions of Vicuna, namely \textbf{Vicuna-13B-v1.5}.

\end{itemize}

\subsection{Implementation Details}\label{sec:implement}

\noindent \textbf{Hyperparameter:} The hyperparameters are all presented in Appendix~\ref{appendix:hyper}.

\noindent \textbf{Retriever:} We adopt BM25 as our retrieval model based on findings from~\cite{ram2023context}, which demonstrated its superior performance in Retrieval-Augmented Generation, even outperforming some dense retrieval models. We also explored the impact of replacing BM25 with a SOTA dense retrieval method SGPT~\cite{muennighoff2022sgpt}, which is detailed in Section~\ref{sec:SGPT}.

\noindent \textbf{Stopwords:} 
For the identification of stopwords within the RIND module, we utilized the en\_core\_web\_sm language model from the Spacy library, a tool recognized for its effectiveness and efficiency in Natural Language Processing tasks as evidenced by previous research~\cite{shelar2020named}.

\noindent \textbf{External Knowledge Corpus:}
We adopt Wikipedia as our external knowledge corpus. Each article are segmented into 100-token passages. 

\noindent \textbf{LLM Configuration}: For the selected LLMs, we directly download model parameters from the official Hugging Face repositories for each model, and use the code provided by Hugging Face to conduct text generation. For the generation configuration, we have chosen greedy decoding as the decoding strategy for LLM inference to ensure the reproducibility of our experimental results. However, for practical applications, we recommend using the official default generation configuration provided by each model, as this will yield better performance.

\section{Experimental Results}

\subsection{Overall Results of DRAGIN and Baselines}
Our experiments comprehensively evaluate the performance of DRAGIN against various baselines across four benchmark datasets: 2WikiMultihopQA, HotpotQA, StrategyQA, and IIRC. The results, summarized in Table~\ref{tab:overall}, underscore several critical insights:
(1) The integration of single-round retrieval augmentation consistently boosts LLMs' performance across all datasets when compared to direct question answering, confirming the effectiveness of RAG.
(2) Despite the overall positive impact of retrieval augmentation, we observe that fixed rules-based retrieval methods, e.g. FL-RAG and FS-RAG, do not always outperform single-round retrieval. This observation validates our hypothesis that retrieval augmentation, if conducted at a wrong position, may not be helpful in improving the quality of LLM's outputs.
This underscores the significance of timing in the activation of retrieval processes, which should be tailored to the information needs of Large Language Models (LLMs), activating retrieval only when LLMs necessitate external knowledge.
(3) The performance of FLARE varies significantly among different datasets. Interestingly, as shown in our ablation study (\S ~\ref{sec:how}), the query formulation strategies are significantly better than those used by other baselines, but its overall performance is not. This indicates that the timing of retrieval augmentation in FLARE is far from perfect.
(4) Our proposed DRAGIN method demonstrates superior performance across most LLMs and datasets. This indicates the robustness and effectiveness of DRAGIN in enhancing LLMs' capabilities.
(5) DRAGIN demonstrates more substantial performance improvements in MultihopQA tasks, such as 2WikiMultihopQA and HotpotQA, than in tasks requiring common sense reasoning, like those in the StrategyQA dataset. This difference highlights DRAGIN's specialized capability in managing complex, multi-step reasoning tasks.

\begin{table}[t]
\caption{Comparison of the frequency of retrieval module activation in dynamic RAG frameworks across all datasets. 2WMQA, HQA, SQA indicates 2WikiMultihopQA, HotpotQA, StrategyQA respectively.}
\centering
\label{tab:efficiency}
\setlength\tabcolsep{4pt} 
\footnotesize
\begin{tabular}{lccccc}
\toprule
                                                            && \textbf{2WMQA} & \textbf{HQA} & \textbf{SQA} & \textbf{IIRC}  \\
                                          \midrule
  &\textbf{} & \textbf{\#Num}           & \textbf{\#Num}    & \textbf{\#Num}      & \textbf{\#Num} \\
 \toprule
                                            \multirow{4}{*}{\textbf{L13B}}&\textbf{FL-RAG} & 3.770                     & 3.194             & 3.626               & 3.426\\
                                            &\textbf{FS-RAG} & 3.131                    & 4.583             & 4.885               & 4.305\\
                                                                                        &\textbf{FLARE}  & 1.592                    & 3.378             & 0.625               & 5.521\\

                                            &\textbf{DRAGIN} & 2.631                    & 3.505             & 4.786               & 2.829\\
                                           \midrule
                                            \multirow{4}{*}{\textbf{L7B}}&\textbf{FL-RAG} & 3.342                    & 3.809             & 3.757               & 2.839\\
                                            &\textbf{FS-RAG} & 3.833                    & 4.152             & 4.546               & 4.210\\
                                                                                        &\textbf{FLARE}  & 0.941                    & 1.064             & 1.271               & 1.095\\

                                            &\textbf{DRAGIN} & 2.836                    & 3.013            & 4.629               & 2.927\\
                                           \midrule
                                            \multirow{4}{*}{\textbf{V13B}}&\textbf{FL-RAG} & 4.199                    & 3.564             & 3.591               & 3.189\\

                                            &\textbf{FS-RAG} & 3.720                     & 5.701             & 6.820& 6.032\\
                                            &\textbf{FLARE}  & 1.093                    & 1.078             & 1.118               & 0.335\\
                                            
                                            &\textbf{DRAGIN} & 2.542                    & 3.184             & 3.744               & 3.120\\
                                          \toprule
\end{tabular}
\end{table}

\subsection{Efficiency}
\label{sec:efficiency}

In this section, we investigate the efficiency of various dynamic RAG frameworks across multiple datasets. We measure efficiency based on the number of retrieval calls made, as outlined in Table~\ref{tab:efficiency}. Due to the special design of FS-RAG, the \#NUM for FS-RAG also indicates the average number of sentences produced by the LLM in response to queries on this dataset. 
Among the evaluated frameworks, FLARE stood out for its efficiency, requiring the fewest retrieval calls across all datasets. DRAGIN followed closely, with fewer retrieval calls than FS-RAG and FL-RAG. 


\subsection{Timing of Retrieval}
\label{sec:when}

In this subsection, we investigate the impact of the timing of retrieval on the performance of dynamic RAG frameworks. Specifically, we fixed the method of query formulation to use the last complete sentence generated by the LLM as the query, and varied the timing of retrieval as the only variable. We examined DRAGIN alongside three existing frameworks: FLARE, FL-RAG, and FS-RAG on the IIRC dataset.
As shown in Table~\ref{tab:when}, DRAGIN consistently outperforms all other dynamic RAG methods. This highlights the effectiveness of our novel approach to determining the optimal moment for retrieval. DRAGIN's superior performance suggests that its method for detecting the real-time information needs of LLMs and triggering retrieval accordingly is particularly adept at enhancing the quality of the generated text.

\begin{table}[]
\caption{The influence of the ‘When to Retrieve’ decision on various dynamic RAG frameworks, with the IIRC dataset as the evaluation benchmark. The best results are in bold. L13B indicates LLaMA2-13B-Chat, V13B indicates Vicuna-13b-v1.5. We fix the query formulation method, the last complete sentence generated by the LLM is selected as the query for all the baselines.}
\label{tab:when}
\centering
\small
\begin{tabular}{ccccc}
\toprule
                               & \textbf{}       & \textbf{EM}    & \textbf{F1}     & \textbf{Prec.}  \\
                               \midrule
\multirow{4}{*}{\textbf{L13B}} & \textbf{FLARE}  & 0.128          & 0.1599          & 0.1677          \\
                               & \textbf{FL-RAG} & 0.155          & 0.1875          & 0.1986          \\
                               & \textbf{FS-RAG} & 0.171          & 0.2061          & 0.2185          \\
                               & \textbf{DRAGIN} & \textbf{0.187} & \textbf{0.2242} & \textbf{0.2319} \\
                               \midrule
\multirow{4}{*}{\textbf{V13B}} & \textbf{FLARE}  & 0.097          & 0.1277          & 0.1324          \\
                               & \textbf{FL-RAG} & 0.099          & 0.1285          & 0.1324          \\
                               & \textbf{FS-RAG} & 0.103          & 0.1344          & 0.1358          \\
                               & \textbf{DRAGIN} & \textbf{0.196} & \textbf{0.2367} & \textbf{0.2476} \\
                               \toprule
\end{tabular}
\end{table}


We also evaluate the impact of varying threshold values within the RIND module on the performance of the DRAGIN framework.
We present the experimental results on the HotpotQA dataset in Table~\ref{tab:threshold}. 
Our experimental results show that DRAGIN's performance remains stable across threshold settings, indicating a low sensitivity to changes in this hyperparameter.
The threshold value is pivotal in determining the retrieval module's activation frequency. 
As the threshold increases, the frequency of the retrieval module's activation decreases, suggesting that adjusting the threshold can strike a balance between the system's efficiency and the accuracy of its outputs in practical applications.

\begin{table}[t]
\caption{Comparasion between different threshold of RIND for LLaMA-13B-Chat model on the HotpotQA dataset. The best results are in bold.}
\label{tab:threshold}
\centering
\small
\setlength\tabcolsep{5pt} 
\begin{tabular}{cccc}
\toprule
\textbf{threshold} & \textbf{EM}    & \textbf{F1}     & \textbf{Prec.}   \\
\midrule
\textbf{0.3}       & 0.295          & 0.3856          & 0.3873          \\
\textbf{0.4}       & 0.297          & 0.387           & 0.389           \\
\textbf{0.5}       & 0.299          & 0.3897          & 0.3915          \\
\textbf{0.6}       & \textbf{0.304} & 0.3931          & \textbf{0.3946} \\
\textbf{0.7}       & \textbf{0.304} & 0.3927          & 0.3937          \\
\textbf{0.8}       & 0.301          & 0.392           & 0.3927          \\
\textbf{0.9}       & 0.301          & \textbf{0.3944} & 0.3947          \\
\textbf{1}         & 0.293          & 0.3869          & 0.3875         \\
\toprule
\end{tabular}
\end{table}

\subsection{Query Formulation}
\label{sec:how}

\begin{table}[t]
\caption{The influence of the query formulation methods on various dynamic RAG frameworks, with the HotpotQA dataset as the evaluation benchmark. The best results are in bold. L13B indicates LLaMA2-13B-Chat, V13B indicates Vicuna-13b-v1.5.}
\label{tab:what}
\centering
\small
\setlength\tabcolsep{5pt} 
\begin{tabular}{ccccc}
\toprule
                               &                       & \textbf{EM} & \textbf{F1} & \textbf{Prec.} \\
                               \toprule
\multirow{5}{*}{\textbf{L13B}} & \textbf{FLARE}        & 0.262       & 0.3674      & 0.3792         \\
                               & \textbf{Full Context} & 0.252       & 0.3584      & 0.3711         \\
                               & \textbf{FS-RAG}       & 0.255       & 0.3574      & 0.3685         \\
                               & \textbf{FL-RAG}       & 0.241       & 0.3394      & 0.3495         \\
                               & \textbf{DRAGIN}       & \textbf{0.314}       & \textbf{0.4238}      & \textbf{0.4401}         \\
                               \midrule
\multirow{5}{*}{\textbf{V13B}} & \textbf{FLARE}        & 0.225       & 0.3366      & 0.3420          \\
                               & \textbf{Full Context} & 0.221       & 0.3402      & 0.3457         \\
                               & \textbf{FS-RAG}       & 0.216       & 0.3432      & 0.3507         \\
                               & \textbf{FL-RAG}       & 0.214       & 0.3268      & 0.3264         \\
                               & \textbf{DRAGIN}       & \textbf{0.288}       & \textbf{0.4164}     &\textbf{0.4226}        \\
                               \toprule
\end{tabular}
\end{table}

This subsection delves into the impact of query formulation techniques on the performance of dynamic RAG frameworks.
We standardize the timing of trigger retrieval to RIND, which is proven to be the most effective timing based on the experimental results detailed in section ~\ref{sec:when}.
We focus on the comparison between DRAGIN and three existing frameworks: FLARE, FL-RAG, and FS-RAG.
The query formulation method of FLARE is the last generated sentence excludes low-probability tokens.
FL-RAG selects the closest 25 tokens to this position as the query.
FS-RAG selects the sentence before this position as the query.
We also evaluate the effectiveness of using the full context as the query.
As shown in Table~\ref{tab:what}, DRAGIN's query formulation method performs best among all the dynamic RAG frameworks.
FLARE emerged as the second most effective query formulation method, outperforming the FS-RAG and FL-RAG methods.
Moreover, leveraging the entire context as a query did not yield optimal results, indicating potential redundancy within the full context. This finding validates the effectiveness of our proposed QFS method, which aims to select tokens from the context that can represent the real-time information needs of the LLM as the query.

\subsection{Impact of Retriever}
\label{sec:SGPT}
In the dynamic RAG paradigm, the choice of the retriever plays an important role in retrieving relevant passages from a corpus based on a given query. 
In the field of information retrieval, the two popular types of retrieval methods are lexical matching~\cite{zhai2008statistical,robertson2009probabilistic} and dense retrieval~\cite{su2023caseformer,gao2021condenser,su2023wikiformer,muennighoff2022sgpt,li2023thuir,ma2023caseencoder,ye2024relevance,su2023thuir2,li2023towards,chen2023thuir,chen2022web,li2023thuir3,fang2024scaling}.
Among lexical matching techniques, BM25 stands out for its widespread adoption and effectiveness~\cite{robertson2009probabilistic}. Conversely, among existing dense retrieval methods, none has achieved the widespread popularity of BM25. We have opted for SGPT, which has recently attained state-of-the-art performance across a variety of datasets.~\cite{muennighoff2022sgpt}.

In our experimental analysis presented in Table \ref{tab:retriever}, we found that BM25 consistently surpasses SGPT in performance across various datasets within the dynamic RAG framework, despite SGPT's generally better performance in numerous information retrieval tasks. This outcome aligns with the findings of prior research, such as the study by~\cite{ram2023context}, which underscored BM25's effectiveness in RAG tasks. These results indicate that despite progress in dense retrieval technologies like SGPT, the simpler, lexicon-based BM25 algorithm is still a strong baseline for enhancing the performance of LLM in RAG tasks.

\begin{table}[t]
\caption{Comparison of performance between BM25 and SGPT using the LLaMA2-13B-Chat model. The method with better performance is highlighted.}
\label{tab:retriever}
\centering
\small
\setlength\tabcolsep{5pt} 
\begin{tabular}{ccccc}
\toprule
                                          & \textbf{retriever} & \textbf{EM}     & \textbf{F1}     & \textbf{Prec.}  \\
                                          \midrule
\multirow{2}{*}{\textbf{2WMQA}} & \textbf{BM25}      & \textbf{0.304}  & \textbf{0.393} & \textbf{0.395} \\
                                          & \textbf{SGPT}      & 0.273           & 0.356          & 0.357          \\
                                          \midrule
\multirow{2}{*}{\textbf{HQA}}        & \textbf{BM25}      & \textbf{0.314}& \textbf{0.424}& \textbf{0.437} \\
                                          & \textbf{SGPT}      & 0.264           & 0.371          & 0.388          \\
                                          \midrule
\multirow{2}{*}{\textbf{IIRC}}            & \textbf{BM25}      & \textbf{0.185} & \textbf{0.222} & \textbf{0.235} \\
                                          & \textbf{SGPT}      & 0.169          & 0.201          & 0.207          \\
                                          \toprule
\end{tabular}
\end{table}



%

\section{Conclusions and Future Works}

In this work, we propose DRAGIN, a dynamic RAG framework tailored to address the real-time information needs of LLMs during text generation. By integrating RIND for timely retrieval activation and QFS for precise query formulation, DRAGIN significantly outperforms existing dynamic RAG methods across various knowledge-intensive benchmarks. 

\section{Limitations}
We acknowledge the limitations of this paper. One of the primary limitations is the reliance on the self-attention mechanism of Transformer-based LLMs for both Real-time Information Needs Detection (RIND) and Query Formulation based on Self-attention (QFS). While self-attention scores are accessible for all open-source LLMs, it's important to note that our method is not applicable to certain APIs that do not provide access to the self-attention scores. Thus, our future work aims to develop more methods to overcome this constraint.

\section{Ethics Statement}
In conducting this research, we have prioritized ethical considerations at every stage to ensure the responsible development and application of AI technologies. 
Our research does not rely on personally identifiable information or require manually annotated datasets. 
We firmly believe in the principles of open research and the scientific value of reproducibility. To this end, we have made all models, data, and code associated with our paper publicly available on GitHub. 
This transparency not only facilitates the verification of our findings by the community but also encourages the application of our methods in other contexts. 

\section*{Acknowledgments}
This work is supported by Quan Cheng Laboratory (Grant No. QCLZD202301).

\bibliography{custom}
\appendix


\section{Datasets and Settings}
\label{appendix:datasets}

Datasets, metrics, and experimental settings are summarized in Table~\ref{tab:settings}.

\begin{itemize}[leftmargin=*]
\item \textbf{2WikiMultihopQA.} For the question "When did the director of film Hypocrite (Film) die?", the output we aim to generate is "The film Hypocrite was directed by Miguel Morayta. Miguel Morayta died on 19 June 2013. So the answer is 19 June 2013." For 2WikiMultihopQA, we employed 6 examples enclosed in ~\cite{trivedi2022interleaving} for context learning, using BM25 as the retriever and Wikipedia articles as the retrieval corpus. While increasing the number of documents can somewhat improve performance, excessive retrieval content may cause the model to overlook previous exemplars. Therefore, we utilized a maximum document count of 3.

\item \textbf{HotpotQA.} For the question "What film directed by Brian Patrick Butler was inspired by a film directed by F.W. Murnau?", the output we aim to generate is "Brian Patrick Butler directed the film The Phantom Hour. The Phantom Hour was inspired by the films such as Nosferatu and The Cabinet of Dr. Caligari. Of these, Nosferatu was directed by F.W. Murnau. So the answer is The Phantom Hour." We utilized 8 examples enclosed in ~\cite{trivedi2022interleaving}, conducted experiments with BM25 as the retriever on the Wikipedia corpus, and retrieved 3 documents for context learning.

\item \textbf{IIRC.} For the question "What is the age difference between the kicker and the quarterback for the Chargers?", the output we aim to generate is "The kicker for the Chargers is Nate Kaeding. The quarterback (QB) for the Chargers is Philip Rivers. Nate Kaeding was born in the year 1982. Philip Rivers was born in the year 1981. So the answer is 1." We utilized 8 examples enclosed in ~\cite{trivedi2022interleaving}, conducted experiments with BM25 as the retriever on the Wikipedia corpus, and retrieved 3 documents for context learning. In particular, we excluded questions without answers, so there are a total of 954 questions in IIRC.

\item \textbf{StrategyQA.} For the question "Is it common to see frost during some college commencements?", the output we aim to generate is "College commencement ceremonies can happen in December, May, and June. December is in the winter, so there can be frost. Thus, there could be frost at some commencements. So the answer is yes." We utilized 8 examples enclosed in \cite{wei2023chainofthought}, conducted experiments with BM25 as the retriever on the Wikipedia corpus, and retrieved 3 documents for context learning.

\end{itemize}

\section{Evaluation Details} 
In order to match the answers obtained by the model, we included the paradigm "So the answer is" in the exemplars to encourage the model to generate in this format. Specifically, if "So the answer is" is absent from all of the model's generations, during the evaluation phase, we append "So the answer is" to the end of the model's output, prompting the model to generate again. Subsequently, we select the words following "So the answer is" as the final answer.

\section{Case Study}
\label{appendix:case}

We select the following question for case study:

\begin{tcolorbox}[colback=lightgray!20,colframe=darkgray!80,title=Question]
The arena where the Lewiston Maineiacs played their home games can seat how many people?
\end{tcolorbox}

This is a complex question that requires identifying the Arena's name as well as finding the seating capacity of the arena.

\begin{tcolorbox}[colback=lightgray!20,colframe=darkgray!80,title=Initial Output of the LLM]
The arena where the Lewiston Maineiacs played their home games is the Androscoggin Bank Colisée. The Androscoggin Bank Colisée has a seating capacity of \textbf{4,250}. Therefore, the answer is 4,250.
\end{tcolorbox}

During the generation process of LLM, the RIND module identified that the LLM needs external knowledge assistance when generating the first \textbf{4,250}, thus triggering the QFS module for query generation. At this moment, the information need of the LLM is to find the seating capacity of Androscoggin Bank Colisée.
Our proposed QFS generates a query based on the self-attention distribution over the entire context, where the tokens selected by QFS are as follows in bold:

\begin{tcolorbox}[colback=lightgray!20,colframe=darkgray!80,title=The selected tokens for query formulation]
Question: The arena where the Lewiston Maineiacs played their home games can \textbf{seat} how many people? </s> Answer: The arena where the Lewiston Maineiacs played their home games is the \textbf{Androscoggin Bank Colisée}. The Androscoggin Bank Colisée has a \textbf{seating capacity} \textbf{of} \textbf{|}
\end{tcolorbox}

Thus, the QFS module generated the following query:

\begin{tcolorbox}[colback=lightgray!20,colframe=darkgray!80,title=The query generated by the QFS module]
seat Androscoggin Bank Colisée seating capacity
\end{tcolorbox}

The generated query indeed reflects the real-time information needs of LLM. This query directly led to the retrieval of the relevant paragraph from Wikipedia:

\begin{tcolorbox}[colback=lightgray!20,colframe=darkgray!80,title=The top 1 retrieved document]
Androscoggin Bank Colisée The Androscoggin Bank Colisée is a 4,000 capacity (3,677 seated) multi-purpose arena, in Lewiston, Maine, that opened in 1958. The Androscoggin Bank Colisée was built to ......
\end{tcolorbox}

Thus, after including the retrieved passage to the LLM, it can then generate the correct answer: 3,677. The final revised output is as follows:

\begin{tcolorbox}[colback=lightgray!20,colframe=darkgray!80,title=Final Output]
The arena where the Lewiston Maineiacs played their home games is the Androscoggin Bank Colisée. The Androscoggin Bank Colisée has a seating capacity of 3,677. Therefore, the answer is 3,677.
\end{tcolorbox}

In contrast, the FLARE framework decides to trigger retrieval for the first sentence because the probability of the tokens `Lewiston', `home', `Androscoggin', and `Bank' are all below the FLARE's threshold:

\begin{tcolorbox}[colback=lightgray!20,colframe=darkgray!80,title=Initial output of the LLM]
The arena where the \sout{Lewiston} Maineiacs played their \sout{home} games is the \sout{Androscoggin} \sout{Bank} Colisée.
\end{tcolorbox}

After that, FLARE generates the following query that removes all the tokens below the threshold from the first sentence:

\begin{tcolorbox}[colback=lightgray!20,colframe=darkgray!80,title=The generated query of FLARE]
The arena where the Maineiacs played their games is the Colisée.
\end{tcolorbox}

Unfortunately, the generated query omitted the correct information from the most recent sentence, leading to an irrelevant passage retrieval for the LLM's real-time information need. Thus, the LLM did not answer this question correctly.

\section{Error Analysis}
\label{appendix:error}
When the RIND module triggers retrieval, the DRAGIN framework adds multiple passages to the input of the LLM, thereby extending the context length. As a result, LLMs that typically perform poorly with long contexts may become confused during the generation process and mix up the information from these passages.

We select the following case:

\begin{tcolorbox}[colback=lightgray!20,colframe=darkgray!80,title=Question]
What is the name of the fight song of the university whose main campus is in Lawrence, Kansas, and whose branch campuses are in the Kansas City metropolitan area?
\end{tcolorbox}

Our framework first detects that the LLM needs to determine which university's main campus is located in Lawrence, Kansas, and thus retrieves three passages:

\begin{tcolorbox}[colback=lightgray!20,colframe=darkgray!80,title=Retrieved Passages]
P1: The Kansas City metropolitan area's largest private employer is Cerner Corporation.....

P2: University of Kansas The University of Kansas, also referred to as KU, is a public research university in the U.S. state of Kansas ......

P3: The population of the Kansas City MSA grew from 1,842,965 to an estimated 2,037,357.....
\end{tcolorbox}

The second of the three passages is relevant. However, the LLaMA-7B model, which performs poorly in long context scenarios, failed to generate the correct answer based on this retrieved external knowledge, despite the passage indeed addressing the LLM's real-time information needs. Therefore, future work should explore how to enable large models to effectively understand complex information in extended contexts.

\section{Hyperparameters}
\label{appendix:hyper}

The hyperparameters of DRAGIN on different datasets are listed in Table~\ref{tab:hyperparameters}.

\begin{table*}[]
\centering
\caption{Dataset statistics and experimental settings.}
\label{tab:settings}
\small
\begin{tabular}{lcccc}
\toprule
           & \multicolumn{4}{c}{\textbf{Dataset statistics}} 
            \\
            \midrule
            & \textbf{2WikiMultihopQA}     & \textbf{HotpotQA}            & \textbf{IIRC}                & \textbf{StrategyQA}     \\ 
            \midrule
Task        & multi-hop QA         & multi-hop QA         & reading comprehension QA      & commonsense QA \\
\#Examples  & 1000                & 1000                & 954                 & 1000           \\ 

\toprule

\\
\\
\\

\toprule
            & \multicolumn{4}{c}{\textbf{Evaluation settings}}                                          \\
            \midrule
& \textbf{2WikiMultihopQA}     & \textbf{HotpotQA}            & \textbf{IIRC}                & \textbf{StrategyQA}     \\ 
\midrule

Metrics     & EM, F1, Prec., Rec. & EM, F1, Prec., Rec. & EM, F1, Prec., Rec. & Accuracy             \\ 
\toprule

\\
\\
\\

\toprule
            & \multicolumn{4}{c}{\textbf{Retrieval settings}}                                           \\
            \midrule
& \textbf{2WikiMultihopQA}     & \textbf{HotpotQA}            & \textbf{IIRC}                & \textbf{StrategyQA}     \\ 
\midrule            
Corpus      & \multicolumn{4}{c}{Wikipedia}                                                    \\
Retriever   & \multicolumn{4}{c}{BM25}                                                         \\
Top-k       & \multicolumn{4}{c}{3}                                                            \\ 
\toprule
\\
\\
\\
\toprule
            & \multicolumn{4}{c}{\textbf{Prompt format}}                                                \\
            \midrule
& \textbf{2WikiMultihopQA}     & \textbf{HotpotQA}            & \textbf{IIRC}                & \textbf{StrategyQA}     \\ 
\midrule            
\#Examplars & 6                   & 8                   & 8                   & 8 \\
\toprule
\end{tabular}
\end{table*}

\begin{table*}[]
\caption{Hyperparameters of DRAGIN on different datasets.}
\label{tab:hyperparameters}
\begin{tabular}{cccccc}
\toprule
\textbf{LLM} &
  \textbf{Hyperparameters} &
  \textbf{2WikiMultihopQA} &
  \textbf{HotpotQA} &
  \textbf{IIRC} &
  \textbf{StrategyQA} \\
\toprule
\multirow{3}{*}{\textbf{Llama2-13b-chat}} &
  generate length &
  64 &
  100 &
  128 &
  100 \\
                                          & $\theta$         & 0.6 & 1.2 & 1.25 & 1.0 \\
                                          & top $n$ tokens    & 25  & 35  & 25   & 25  \\
\midrule
\multirow{3}{*}{\textbf{Llama2-7b-chat}}  & generate length & 64  & 100 & 128  & 100          \\
                                          & $\theta$         & 1.0 & 1.3 & 1.3  & 0.75         \\
                                          & top $n$ tokens    & 25  & 35  & 35   & 35           \\
\midrule
\multirow{3}{*}{\textbf{Vicuna-13b-v1.5}} & generate length & 64  & 100 & 128  & 100          \\
                                          & $\theta$         & 1.2 & 1.2 & 1.3  & 1.5          \\
                                          & top $n$ tokens    & 25  & 35  & 35   & 25          \\
\toprule
\end{tabular}
\end{table*}

\clearpage

\section{Prompt Template}
\label{appendix:prompt}

Each dataset has a prompt for direct generation and a prompt for generation with relevant documents, as shown below.

    
\begin{tcolorbox}[colback=lightgray!20,colframe=darkgray!80,title=Prompt 1: exemplars of 2WMQA Direct]
\small
Question: When did the director of film Hypocrite (Film) die?

Answer: The film Hypocrite was directed by Miguel Morayta. Miguel Morayta died on 19 June 2013. So the answer is 19 June 2013.

Question: Are both Kurram Garhi and Trojkrsti located in the same country?

Answer: Kurram Garhi is located in the country of Pakistan. Trojkrsti is located in the country of Republic of Macedonia. Thus, they are not in the same country. So the answer is no.

Question: Do director of film Coolie No. 1 (1995 Film) and director of film The Sensational Trial have the same nationality?

Answer: Coolie No. 1 (1995 film) was directed by David Dhawan. The Sensational Trial was directed by Karl Freund. David Dhawan's nationality is India. Karl Freund's nationality is Germany. Thus, they do not have the same nationality. So the answer is no.

Question: Who is Boraqchin (Wife Of Ögedei)'s father-in-law?

Answer: Boraqchin is married to Ögedei Khan. Ögedei Khan's father is Genghis Khan. Thus, Boraqchin's father-in-law is Genghis Khan. So the answer is Genghis Khan.

Question: Who was born first out of Martin Hodge and Ivania Martinich?

Answer: Martin Hodge was born on 4 February 1959. Ivania Martinich was born on 25 July 1995. Thus, Martin Hodge was born first. So the answer is Martin Hodge.

Question: When did the director of film Laughter In Hell die?

Answer: The film Laughter In Hell was directed by Edward L. Cahn. Edward L. Cahn died on August 25, 1963. So the answer is August 25, 1963.

Question: Who is the mother of the director of film Polish-Russian War (Film)?

Answer:
\end{tcolorbox}


\begin{figure*}
    
\begin{tcolorbox}[colback=lightgray!20,colframe=darkgray!80,title=Prompt 2: exemplars of 2WMQA RAG]
\small

Question: When did the director of film Hypocrite (Film) die?

Answer: The film Hypocrite was directed by Miguel Morayta. Miguel Morayta died on 19 June 2013. So the answer is 19 June 2013.

Question: Are both Kurram Garhi and Trojkrsti located in the same country?

Answer: Kurram Garhi is located in the country of Pakistan. Trojkrsti is located in the country of Republic of Macedonia. Thus, they are not in the same country. So the answer is no.

Question: Do director of film Coolie No. 1 (1995 Film) and director of film The Sensational Trial have the same nationality?

Answer: Coolie No. 1 (1995 film) was directed by David Dhawan. The Sensational Trial was directed by Karl Freund. David Dhawan's nationality is India. Karl Freund's nationality is Germany. Thus, they do not have the same nationality. So the answer is no.

Question: Who is Boraqchin (Wife Of Ögedei)'s father-in-law?

Answer: Boraqchin is married to Ögedei Khan. Ögedei Khan's father is Genghis Khan. Thus, Boraqchin's father-in-law is Genghis Khan. So the answer is Genghis Khan.

Question: Who was born first out of Martin Hodge and Ivania Martinich?

Answer: Martin Hodge was born on 4 February 1959. Ivania Martinich was born on 25 July 1995. Thus, Martin Hodge was born first. So the answer is Martin Hodge.

Question: When did the director of film Laughter In Hell die?

Answer: The film Laughter In Hell was directed by Edward L. Cahn. Edward L. Cahn died on August 25, 1963. So the answer is August 25, 1963.

Context:

[1] film was shot between May 6 and 18 June 2008 in locations of Warsaw, Wejherowo, Sopot and Gdynia outskirts. The film premiered on May 22, 2009. The budget of Polish-Russian War amounted to approx. 4 million zlotys. The creators of the music for the film are Jan Komar, Filip Kuncewicz, Liroy, Mateusz Łapot and Jarosław Karczmarczyk. The soundtrack also included the following songs: Polish-Russian War (film) Polish-Russian War (Wojna polsko-ruska) is a 2009 Polish film directed by Xawery Żuławski based on the novel Polish-Russian War under the white-red flag by Dorota Masłowska. 

[2] Pharaoh (film) Pharaoh () is a 1966 Polish film directed by Jerzy Kawalerowicz and adapted from the eponymous novel by the Polish writer Bolesław Prus. In 1967 it was nominated for an Academy Award for Best Foreign Language Film. It was also entered into the 1966 Cannes Film Festival. Jerzy Kawalerowicz, who had previously directed such films as "Cellulose" (1953), "Under the Phrygian Star" (1954), "The Shade" (1956), "The Real End of the Great War" (1957), "Night Train" (1959) and "Mother Joan of the Angels" (1961)

[3] Polish-Russian War (film) Polish-Russian War (Wojna polsko-ruska) is a 2009 Polish film directed by Xawery Żuławski based on the novel Polish-Russian War under the white-red flag by Dorota Masłowska. The film's events take place over several days and they are set in the present time in a large Polish city. The main character is a bandit, a Polish dres (a Polish chav) called "Strong" (Borys Szyc), who does not work or study, and who frequently gets into conflict with the law and is in love with Magda (Roma Gąsiorowska). The relationship is not going well. 

\vspace{10mm}

Answer in the same format as before.

Question: Who is the mother of the director of film Polish-Russian War (Film)?

\end{tcolorbox}

\end{figure*}

\begin{figure*}
    
\begin{tcolorbox}[colback=lightgray!20,colframe=darkgray!80,title=Prompt 3: exemplars of HotpotQA Direct]
\small
Question: Jeremy Theobald and Christopher Nolan share what profession?

Answer: Jeremy Theobald is an actor and producer. Christopher Nolan is a director, producer, and screenwriter. Therefore, they both share the profession of being a producer. So the answer is producer.

Question: What film directed by Brian Patrick Butler was inspired by a film directed by F.W. Murnau?

Answer: Brian Patrick Butler directed the film The Phantom Hour. The Phantom Hour was inspired by the films such as Nosferatu and The Cabinet of Dr. Caligari. Of these Nosferatu was directed by F.W. Murnau. So the answer is The Phantom Hour.

Question: How many episodes were in the South Korean television series in which Ryu Hye-young played Bo-ra?

Answer: The South Korean television series in which Ryu Hye-young played Bo-ra is Reply 1988. The number of episodes Reply 1988 has is 20. So the answer is 20.

Question: Were Lonny and Allure both founded in the 1990s?

Answer: Lonny (magazine) was founded in 2009. Allure (magazine) was founded in 1991. Thus, of the two, only Allure was founded in 1990s. So the answer is no.

Question: Vertical Limit stars which actor who also played astronaut Alan Shepard in "The Right Stuff"?

Answer: The actor who played astronaut Alan Shepard in "The Right Stuff" is Scott Glenn. The movie Vertical Limit also starred Scott Glenn. So the answer is Scott Glenn.

Question: What was the 2014 population of the city where Lake Wales Medical Center is located?

Answer: Lake Wales Medical Center is located in the city of Polk County, Florida. The population of Polk County in 2014 was 15,140. So the answer is 15,140.

Question: Who was born first? Jan de Bont or Raoul Walsh?

Answer: Jan de Bont was born on 22 October 1943. Raoul Walsh was born on March 11, 1887. Thus, Raoul Walsh was born the first. So the answer is Raoul Walsh.

Question: In what country was Lost Gravity manufactured?

Answer: The Lost Gravity (roller coaster) was manufactured by Mack Rides. Mack Rides is a German company. So the answer is Germany.

\vspace{10mm}

Answer the following question by reasoning step-by-step, following the example above.

Question: Were Scott Derrickson and Ed Wood of the same nationality?

\end{tcolorbox}
\end{figure*}

\begin{figure*}
    
\begin{tcolorbox}[colback=lightgray!20,colframe=darkgray!80,title=Prompt 4: exemplars of HotpotQA RAG]
\small

Question: Jeremy Theobald and Christopher Nolan share what profession?

Answer: Jeremy Theobald is an actor and producer. Christopher Nolan is a director, producer, and screenwriter. Therefore, they both share the profession of being a producer. So the answer is producer.

Question: What film directed by Brian Patrick Butler was inspired by a film directed by F.W. Murnau?

Answer: Brian Patrick Butler directed the film The Phantom Hour. The Phantom Hour was inspired by the films such as Nosferatu and The Cabinet of Dr. Caligari. Of these Nosferatu was directed by F.W. Murnau. So the answer is The Phantom Hour.

Question: How many episodes were in the South Korean television series in which Ryu Hye-young played Bo-ra?

Answer: The South Korean television series in which Ryu Hye-young played Bo-ra is Reply 1988. The number of episodes Reply 1988 has is 20. So the answer is 20.

Question: Were Lonny and Allure both founded in the 1990s?

Answer: Lonny (magazine) was founded in 2009. Allure (magazine) was founded in 1991. Thus, of the two, only Allure was founded in 1990s. So the answer is no.

Question: Vertical Limit stars which actor who also played astronaut Alan Shepard in "The Right Stuff"?

Answer: The actor who played astronaut Alan Shepard in "The Right Stuff" is Scott Glenn. The movie Vertical Limit also starred Scott Glenn. So the answer is Scott Glenn.

Question: What was the 2014 population of the city where Lake Wales Medical Center is located?

Answer: Lake Wales Medical Center is located in the city of Polk County, Florida. The population of Polk County in 2014 was 15,140. So the answer is 15,140.

Question: Who was born first? Jan de Bont or Raoul Walsh?

Answer: Jan de Bont was born on 22 October 1943. Raoul Walsh was born on March 11, 1887. Thus, Raoul Walsh was born the first. So the answer is Raoul Walsh.

Question: In what country was Lost Gravity manufactured?

Answer: The Lost Gravity (roller coaster) was manufactured by Mack Rides. Mack Rides is a German company. So the answer is Germany.

Context:

[1] Scott Derrickson Scott Derrickson (born July 16, 1966) is an American director, screenwriter and producer. He lives in Los Angeles, California. Derrickson is best known for directing numerous horror films, such as "The Exorcism of Emily Rose" (2005), "Sinister" (2012), and "Deliver Us From Evil" (2014), as well as the Marvel Cinematic Universe superhero film "Doctor Strange" (2016). Derrickson grew up in Denver, Colorado.

[2] Scott Derrickson Scott Derrickson (born July 16, 1966) is an American director, screenwriter and producer. He lives in Los Angeles, California. Derrickson is best known for directing numerous horror films, such as "The Exorcism of Emily Rose" (2005), "Sinister" (2012), and "Deliver Us From Evil" (2014), as well as the Marvel Cinematic Universe superhero film "Doctor Strange" (2016). Derrickson grew up in Denver, Colorado. He graduated from Biola University with a B.A. in Humanities, with an emphasis on literature and philosophy, and a B.A. in communications, with an emphasis on film, and a minor in theological studies.

[3] The film had its world premiere at the 2013 Toronto International Film Festival. It was released in 2014. Derrickson directed his own script, "Deliver Us from Evil", for producer Jerry Bruckheimer and Sony Screen Gems. Eric Bana played the lead role, and the film was released wide in theaters on July 2, 2014. In 2014, Derrickson wrote a film version of "The Outer Limits" with Cargill. Other upcoming Derrickson projects include an adaptation of Stephen King's "The Breathing Method" with Jason Blum producing, and an adaptation of the popular video game "" for CBS Films.

\vspace{10mm}

Answer in the same format as before.

Answer the following question by reasoning step-by-step, following the example above.

Question: Were Scott Derrickson and Ed Wood of the same nationality?
\end{tcolorbox}
\end{figure*}

\begin{figure*}
\begin{tcolorbox}[colback=lightgray!20,colframe=darkgray!80,title=Prompt 5: exemplars of StrategyQA Direct]
\small
Question: Do hamsters provide food for any animals?

Answer: Hamsters are prey animals. Prey are food for predators. Thus, hamsters provide food for some animals. So the answer is yes.

Question: Could Brooke Shields succeed at University of Pennsylvania?

Answer: Brooke Shields went to Princeton University. Princeton University is about as academically rigorous as the University of Pennsylvania. Thus, Brooke Shields could also succeed at the University of Pennsylvania. So the answer is yes.

Question: Hydrogen's atomic number squared exceeds number of Spice Girls?

Answer: Hydrogen has an atomic number of 1. 1 squared is 1. There are 5 Spice Girls. Thus, Hydrogen's atomic number squared is less than 5. So the answer is no.

Question: Is it common to see frost during some college commencements?

Answer: College commencement ceremonies can happen in December, May, and June. December is in the winter, so there can be frost. Thus, there could be frost at some commencements. So the answer is yes.

Question: Could a llama birth twice during War in Vietnam (1945-46)?

Answer: The War in Vietnam was 6 months. The gestation period for a llama is 11 months, which is more than 6 months. Thus, a llama could not give birth twice during the War in Vietnam. So the answer is no.

Question: Would a pear sink in water?

Answer: The density of a pear is about $0.6g/cm^3$, which is less than water. Objects less dense than water float. Thus, a pear would float. So the answer is no.

\vspace{10mm}

{Following the examples above, answer the question by reasoning step-by-step.}

{Question: Are more people today related to Genghis Khan than Julius Caesar?}
\end{tcolorbox}
\end{figure*}

\begin{figure*}
\begin{tcolorbox}[colback=lightgray!20,colframe=darkgray!80,title=Prompt 6: exemplars of StrageQA RAG]
\small
Question: Do hamsters provide food for any animals?

Answer: Hamsters are prey animals. Prey are food for predators. Thus, hamsters provide food for some animals. So the answer is yes.

Question: Could Brooke Shields succeed at University of Pennsylvania?

Answer: Brooke Shields went to Princeton University. Princeton University is about as academically rigorous as the University of Pennsylvania. Thus, Brooke Shields could also succeed at the University of Pennsylvania. So the answer is yes.

Question: Hydrogen's atomic number squared exceeds number of Spice Girls?

Answer: Hydrogen has an atomic number of 1. 1 squared is 1. There are 5 Spice Girls. Thus, Hydrogen's atomic number squared is less than 5. So the answer is no.

Question: Is it common to see frost during some college commencements?

Answer: College commencement ceremonies can happen in December, May, and June. December is in the winter, so there can be frost. Thus, there could be frost at some commencements. So the answer is yes.

Question: Could a llama birth twice during War in Vietnam (1945-46)?

Answer: The War in Vietnam was 6 months. The gestation period for a llama is 11 months, which is more than 6 months. Thus, a llama could not give birth twice during the War in Vietnam. So the answer is no.

Question: Would a pear sink in water?

Answer: The density of a pear is about $0.6g/cm^3$, which is less than water. Objects less dense than water float. Thus, a pear would float. So the answer is no.

Context:

[1] Borte Chino (Grey Wolf) and his wife was Gua Maral (White Doe) Temujin (Genghis Khan) - Founder and Khagan of the Mongol Empire (1206–1227) Family tree of Genghis Khan The family tree of Genghis Khan is listed below. This family tree only lists prominent members of the Borjigin family and does not reach the present. Genghis Khan appears in the middle of the tree, and Kublai Khan appears at the bottom of the tree. The Borjigin family was the royal family of the Mongol Empire, dating back to the 13th and 14th centuries.

[2] Genghis Khan (2004 TV series) Genghis Khan is a Chinese television series based on the life of Genghis Khan, the founder of the Mongol Empire in the 13th century. Ba Sen, who portrayed the eponymous character in the series, is a descendant of Genghis Khan's second son, Chagatai. The series was first broadcast on CCTV in China in 2004, on KBS in South Korea in 2005, and the Turkish state channel TRT 1. The 30 episodes long television series depicts the major events in the life of Temüjin, the founder of the Mongol Empire in the 13th century. 

[3] He allegedly planned to assassinate Genghis Khan. Although Toghrul was allegedly saved on multiple occasions by Genghis Khan, he gave in to his son and became uncooperative with Genghis Khan. Genghis Khan learned of Senggum's intentions and eventually defeated him and his loyalists. One of the later ruptures between Genghis Khan and Toghrul was Toghrul's refusal to give his daughter in marriage to Jochi, Genghis Khan's first son. This was disrespectful in Mongolian culture and led to a war. 

\vspace{10mm}

Answer in the same format as before.

Following the examples above, answer the question by reasoning step-by-step.

Question: Are more people today related to Genghis Khan than Julius Caesar?

\end{tcolorbox}
\end{figure*}

\begin{figure*}
\begin{tcolorbox}[colback=lightgray!20,colframe=darkgray!80,title=Prompt 7: exemplars of IIRC Direct]
\small
Question: What is the age difference between the kicker and the quarterback for the Chargers?

Answer: The kicker for the Chargers is Nate Kaeding. The quarterback (QB) for the Chargers is Philip Rivers. Nate Kaeding was born in the year 1982. Philip Rivers was born in the year 1981. Thus, the age difference between them is of 1 year. So the answer is 1.

Question: How many years was the ship that took the battalion from New South Wales to Ceylon in service?

Answer: The ship that took the battalion from New South Wales to Ceylon is General Hewitt. General Hewitt was launched in Calcutta in 1811. General Hewitt was sold for a hulk or to be broken up in 1864. So she served for a total of 1864 - 1811 = 53 years. So the answer is 53.

Question: What year was the theatre that held the 2016 NFL Draft built?

Answer: The theatre that held the 2016 NFL Draft is Auditorium Theatre. The Auditorium Theatre was built in 1889. So the answer is 1889.

Question: How long had Milan been established by the year that Nava returned there as a reserve in the first team's defense?

Answer: Nava returned to Milan as a reserve in the first team's defense in the year 1990. Milan had been established in the year 1899. Thus, Milan had been established for 1990 - 1899 = 91 years when Milan returned to Milan as a reserve in the first team's defense. So the answer is 91.

Question: When was the town Scott was born in founded?

Answer: Scott was born in the town of Cooksville, Illinois. Cooksville was founded in the year 1882. So the answer is 1882.

Question: In what country did Wright leave the French privateers?

Answer: Wright left the French privateers in Bluefield's river. Bluefields is the capital of the South Caribbean Autonomous Region (RAAS) in the country of Nicaragua. So the answer is Nicaragua.

Question: Who plays the A-Team character that Dr. Hibbert fashioned his hair after?

Answer: Dr. Hibbert fashioned his hair after Mr. T from The A-Team. Mr T.'s birthname is Lawrence Tureaud. So the answer is Lawrence Tureaud.

Question: How many people attended the conference held near Berlin in January 1942?

Answer: The conference held near Berlin in January 1942 is Wannsee Conference. Wannsee Conference was attended by 15 people. So the answer is 15.

Question: In what country did Bain attend doctoral seminars of Wlad Godzich?

Answer:
\end{tcolorbox}
\end{figure*}

\begin{figure*}
\begin{tcolorbox}[colback=lightgray!20,colframe=darkgray!80,title=Prompt 8: exemplars of IIRC RAG]
\small
Question: What is the age difference between the kicker and the quarterback for the Chargers?

Answer: The kicker for the Chargers is Nate Kaeding. The quarterback (QB) for the Chargers is Philip Rivers. Nate Kaeding was born in the year 1982. Philip Rivers was born in the year 1981. Thus, the age difference between them is of 1 year. So the answer is 1.

Question: How many years was the ship that took the battalion from New South Wales to Ceylon in service?

Answer: The ship that took the battalion from New South Wales to Ceylon is General Hewitt. General Hewitt was launched in Calcutta in 1811. General Hewitt was sold for a hulk or to be broken up in 1864. So she served for a total of 1864 - 1811 = 53 years. So the answer is 53.

Question: What year was the theatre that held the 2016 NFL Draft built?

Answer: The theatre that held the 2016 NFL Draft is Auditorium Theatre. The Auditorium Theatre was built in 1889. So the answer is 1889.

Question: How long had Milan been established by the year that Nava returned there as a reserve in the first team's defense?

Answer: Nava returned to Milan as a reserve in the first team's defense in the year 1990. Milan had been established in the year 1899. Thus, Milan had been established for 1990 - 1899 = 91 years when Milan returned to Milan as a reserve in the first team's defense. So the answer is 91.

Question: When was the town Scott was born in founded?

Answer: Scott was born in the town of Cooksville, Illinois. Cooksville was founded in the year 1882. So the answer is 1882.

Question: In what country did Wright leave the French privateers?

Answer: Wright left the French privateers in Bluefield's river. Bluefields is the capital of the South Caribbean Autonomous Region (RAAS) in the country of Nicaragua. So the answer is Nicaragua.

Question: Who plays the A-Team character that Dr. Hibbert fashioned his hair after?

Answer: Dr. Hibbert fashioned his hair after Mr. T from The A-Team. Mr T.'s birthname is Lawrence Tureaud. So the answer is Lawrence Tureaud.

Question: How many people attended the conference held near Berlin in January 1942?

Answer: The conference held near Berlin in January 1942 is Wannsee Conference. Wannsee Conference was attended by 15 people. So the answer is 15.

Context:

[1] Wlad Godzich Wlad Godzich (born May 13, 1945 in Germany, raised in France) is a literary critic, literary theorist, translator, and scholar. He is attributed with influencing the conceptualization of modern literary critical theory. He currently serves as Professor of general and comparative literature, and critical studies at the University of California, Santa Cruz. Godzich has published and translated several books, edited eight collections of essays, and authored over a hundred scholarly articles, lectures, and papers. In 2000, Godzich joined the University of California, Santa Cruz as dean of Humanities. 

[2] The patterns of his thought emerge from his interest in the relationship between language and literacy-the latter conceived as "a determinate set of relations that we have to language." Godzich reinvigorates the semiological project proposed by Saussure but forsaken by his heirs: that of exploring the social functioning of language in its historical and rhetorical actualizations. Wlad Godzich Wlad Godzich (born May 13, 1945 in Germany, raised in France) is a literary critic, literary theorist, translator, and scholar. He is attributed with influencing the conceptualization of modern literary critical theory.

[3] He sits on the editorial board of multiple American, European and Asian journals, both print and electronic. His research grants have been primarily from US, Canadian, Swedish, Swiss and private agencies. Through his work at the University of Minnesota Press, Godzich brought important works of critical theory into English translation. His essays during this period were well received by critics as they were among the first to link deconstruction, cultural criticism, and third-world literatures through linguistics: they can now be seen as tesserae composing a theoretical mosaic of remarkable scope.

\vspace{10mm}

Answer in the same format as before.

Question: In what country did Bain attend doctoral seminars of Wlad Godzich?

Answer: 
\end{tcolorbox}
\end{figure*}

\end{document}